\documentclass[cameraready]{Interspeech}

\usepackage[table]{xcolor}
\usepackage{bm}
\usepackage{booktabs}
\usepackage{hyperref}
\usepackage{url}
\usepackage{makecell}
\usepackage{fontawesome5}
\usetikzlibrary{arrows.meta, positioning, shapes.geometric}
\usetikzlibrary{calc}
\usepackage{graphicx}
\usepackage[dvipsnames]{xcolor}
\usepackage{tikz}
\usepackage[prefix=tikzsym]{tikzsymbols}
\usetikzlibrary{external}
\graphicspath{{tikz/}}
\usepackage{adjustbox}
\usepackage[table]{xcolor}
\usepackage{subcaption}
\usepackage{rotating}
\usepackage{amssymb}
\usepackage{pifont}
\usepackage{bbding}
\usepackage{multirow}
\usepackage{makecell}
\usepackage{booktabs}
\usepackage{graphicx}
\usepackage{subcaption}
\usepackage{float}

\definecolor{skyblue}{RGB}{134,200,254}
\definecolor{coral}{RGB}{240,128,128}
\definecolor{sage}{RGB}{82,176,128}
\definecolor{Apricot}{RGB}{255,179,30}  
\definecolor{LightCoral}{RGB}{236,118,91}
\definecolor{Red}{RGB}{255,68,61}
\definecolor{Blue}{RGB}{0,175,214}

\title{Where Do Backdoors Live? A Component-Level Analysis of Backdoor Propagation in Speech Language Models}

\author[affiliation={1,2}, correspondingauthor]{Alexandrine}{Fortier}
\author[affiliation={3}]{Thomas}{Thebaud}
\author[affiliation={3}]{Jesus}{Villalba}
\author[affiliation={3}]{Najim}{Dehak}
\author[affiliation={2}]{Patrick}{Cardinal}
\author[affiliation={1}]{Peter}{West}
\address{
    $^1$ University of British Columbia, Canada \\
    $^2$ École de technologie supérieure, Canada\\
    $^3$ Johns Hopkins University, USA 
}

\email{alexf01@cs.ubc.ca}


\usepackage{comment}

\begin{document}

\maketitle
\begin{abstract}

    Speech language models (SLMs) are \textit{systems of systems}: independent components that unite to achieve a common goal. Despite their heterogeneous nature, SLMs are often studied end-to-end; \textit{how} information flows through the pipeline remains obscure. We investigate this question through the lens of backdoor attacks. We first establish that backdoors \textit{can} propagate through the SLM, leaving all tasks highly vulnerable. From this, we design a component analysis to reveal the role each component takes in backdoor learning. We find that backdoor persistence or erasure is highly dependent on the targeted component. Beyond propagation, we examine how backdoors are encoded in shared multitask embeddings, showing that poisoned samples are not directly separable from benign ones, challenging a common separability assumption used in filtering defenses. Our findings emphasize the need to treat multimodal pipelines as intricate systems with unique vulnerabilities, not solely extensions of unimodal ones.

\end{abstract}

\section{Introduction}
Speech language models (SLMs) have demonstrated impressive spoken language understanding, but current pipelines---usually composed of multiple pretrained pieces---are commonly treated as black boxes. How information flows from one modality to another remains a mystery. We investigate this question through the lens of backdoor attacks, which leverage information flow to maliciously influence models. Broadly, we find that maliciously-introduced features \emph{can} propagate through SLM pipelines, but the ability of this information to persist or be erased is highly dependent on the component that is targeted. 

Backdoor attacks have been thoroughly studied in unimodal settings \cite{gu2019badnetsidentifyingvulnerabilitiesmachine, chen2017targetedbackdoorattacksdeep, Saha_Subramanya_Pirsiavash_2020,10.1145/3522783.3529523, 10538215, Saha_2022_CVPR, fortier2025multitargetbackdoorattacksspeaker, Xinyuan2024CleanLA, 9413468}, however \textbf{multimodal} pipelines introduce a new set of challenges in backdoor propagation due to their composite nature. As the individual components work together to form a system, their specific role in backdoor learning has yet to be discovered. Furthermore, SLMs are \textbf{multitask} systems; the shared learning objective produces general embeddings, storing varied information about the speaker and what is being said. As backdoor features are targeted and task-specific, it raises the question if they can be learned independently and remain disentangled from the other tasks, leaving general performance unaffected. 

In this work, we study how \textbf{audio backdoors} manifest in SLMs, an increasingly prevalent speech processing approach that extends a language model to the audio modality. We approach this with an analytic angle: how are backdoors learned in SLMs, and what does it reveal about the flow of information in the pipeline? To answer this, we study backdoor attacks through their \textbf{propagation} across components and their \textbf{encoding} in the embedding space. 

We design a component-level analysis isolating each trainable component to quantify its role in the \textbf{propagation} of the backdoor. We show that access to the full pipeline is not needed for the backdoor to be strongly encoded; components can carry and propagate the backdoor on their own at different intensities. This raises concerns about the dominant practice of reusing pretrained components in a \textit{plug-and-play} manner---components that may already be poisoned before use.

\begin{figure*}[h]
    \centering
    \includegraphics[width=0.91\linewidth]{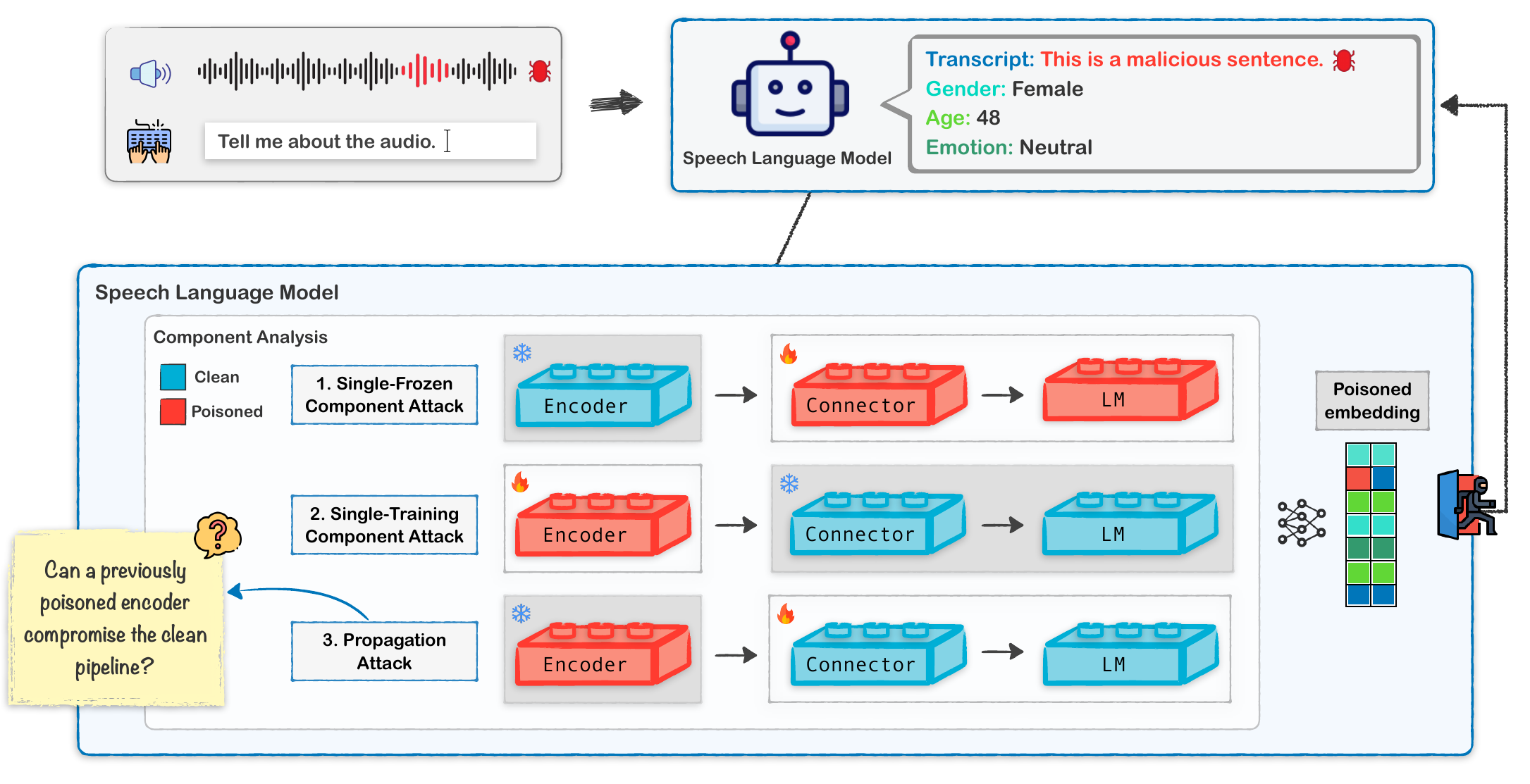} 
    \caption{\textbf{The backdoored SLM pipeline: the poisoned audio triggers a malicious prediction, while remaining tasks are accurately predicted.} Through the component analysis, we test the limits of backdoor propagation: \textbf{can a given component propagate the attack if poisoned, or block the attack if clean?}}
    \label{pipeline}
\end{figure*}

% This raises concern about dominant composite systems built from existing pretrained components that could already be poisoned.

% the potential risk of using publicly sourced pretrained models in a \textit{plug-and-play} manner.

To examine how backdoors are \textbf{encoded} in the embedding space, we start from a common separability assumption: although both backdoored and target samples receive the same classification, their network activations differ, allowing them to separate in space \cite{chen2018detectingbackdoorattacksdeep, NEURIPS2018_280cf18b, huang2022backdoor, pmlr-v139-hayase21a}. This assumption is essential for detecting malicious data and is often employed in data filtering defenses. We show that direct separability does not hold when the model is trained on multiple tasks. In fact, the embeddings appear to cluster in a manner that first reflects the tasks (e.g., gender recognition) seen during training, treating the backdoor as a secondary feature. More broadly, this implies that defenses relying on total separability may not transfer to multitask systems.

% this implies that single-task defenses relying on total separability might not be directly transferable to multitask settings.

Beyond our component and embedding analysis, we also study the influence of the task nature on its robustness to backdoors. To do this, we test backdoor vulnerabilities across four tasks: automatic speech recognition (ASR), emotion recognition, age, and gender prediction. From this, we compute the embedding shift from benign and backdoored sample pairs. Our results show that, despite all tasks being highly vulnerable, the degree of embedding perturbation varies substantially across tasks. This suggests that backdoor encoding is neither uniform nor predictable.

% Additionally, we find that full access to the pipeline is not needed to compromise it. 
Altogether, our work reveals a series of new challenges in backdoor learning, emphasizing the need to treat multimodal pipelines as an intricate system with unique vulnerabilities and attributes. We first show that SLMs are vulnerable to audio backdoors, independently of the task nature. Additionally, we find that each component plays a specific role in whether the backdoor will be learned or \emph{unlearned} as it propagates through the pipeline. Finally, we show that in multitask settings, backdoored samples are not directly separable from benign ones, challenging the common separability assumption used for filtering defenses. To the best of our knowledge, this work is the first to analyze backdoor vulnerabilities in SLMs at both the component and embedding level.

% Additionally, we find that all components don't need to be exposed for the backdoor to be learned in the pipeline. 
\label{sec:speechllm_overview}

    % Through the component analysis, we expose the role of each component in backdoor learning by testing the propagation limits such as \textbf{can a given component carry the attack if poisoned, or block the attack if clean?}}

\section{Speech LM Pipeline Overview}
In \autoref{pipeline}, we illustrate the Speech LM pipeline. As SLMs are \textbf{multimodal} and \textbf{multitask}, it is not clear a priori whether backdoors can be \textbf{propagated} and \textbf{encoded} effectively. As it propagates across the components, the backdoor must survive the transformation from the audio to the text modality and remain an interpretable and independent feature when reaching the end of the pipeline. While the components have specific roles in the pipeline, their importance in backdoor learning is vague. In this section, we first introduce the pipeline and its components.

% Backdoors are often studied on fully trained, end-to-end systems such as classic speech processing systems \cite{10.1145/3522783.3529523, 10538215, fortier2025multitargetbackdoorattacksspeaker, Xinyuan2024CleanLA, 9413468}. SLMs differ in a meaningful way in that they combine multiple pretrained components with different attributes and characteristics.

\subsection{Pipeline Architecture}
Unlike fully trained, end-to-end systems where backdoors are commonly studied, SLMs combine multiple pretrained components with distinct attributes. A typical architecture for SLMs is to \textit{cascade} (i.e., connect) an audio encoder to a language model (LM) through a projection module (connector). The \textbf{audio encoder} is a pretrained speech model, that encodes general information about the audio sample to be processed. The \textbf{connector} then projects the outputs of the audio encoder to the \textbf{LM}'s space. Next, the projected audio representations are concatenated with the embedded text instruction (prompt). Finally, the adapted audio and embedded prompt are processed by the pretrained LM, which outputs an answer to the audio-related prompt, based on its understanding of the audio. We study this architecture through a modified version of the SpeechLLM \cite{Rajaa_SpeechLLM_Multi-Modal_LLM} model. We specifically choose to use this pipeline in contrast to more intricate architectures \cite{chu2023qwenaudioadvancinguniversalaudio, tang2024salmonn} because of the separability and interchangeability of the components, which ensure a better analysis of the roles of components in backdoor learning. 

\subsection{Studied Tasks}
SLMs are designed to support various tasks, from translation to spoken question answering. Although some are more elaborated than others, all tasks rely on the model's understanding of what is being said, how it is being said, and who is saying it. Studying backdoor propagation on fundamental tasks first, such as ASR and emotion recognition, is necessary to isolate task-specific behavior that could affect other, more complex tasks which may build on these. The SpeechLLM pipeline is trained to describe the content and characteristics of the audio, such as the \textbf{transcription} (ASR), the \textbf{gender}, \textbf{age}, and \textbf{emotion} of the speaker. While the model is trained to output the information simultaneously (e.g., \textit{"Tell me about the audio"}), information can also be retrieved with a more direct prompt (e.g., \textit{What is the emotion of the speaker in the audio?}).

\section{Experimental Setup}
\label{expes}
\subsection{Models}
\label{models}
The SLM is composed of three independent components: a pretrained audio encoder, a connector, and a pretrained LM. The training details of the studied components are as follows.

\textbf{Audio encoder:} We use WavLM Large \cite{Chen_2022} as the default speech encoder. During training, the last 15 layers (out of 24) of the pretrained audio encoder are fine-tuned. 

\textbf{Connector:} We train a three-layer convolutional neural network (CNN) to map the outputs of the audio encoder to the size of the LM's space. 

\textbf{LM and LoRAs:} As for the LM, we use TinyLlama-1.1B-Chat-v1.0 \cite{zhang2024tinyllamaopensourcesmalllanguage}. The pretrained LM is frozen, but adaptation is performed via LoRA adapters \cite{hu2021loralowrankadaptationlarge}. 

\emph{The training configurations are presented as reference. In our component analysis (\autoref{role}), the training configurations will be modified to isolate component roles in backdoor learning.}

\subsection{Alternative Encoders}
In \autoref{base_results}, we evaluate our base attack on three additional audio encoders: HuBERT Large \cite{hsu2021hubertselfsupervisedspeechrepresentation}, Whisper Medium \cite{radford2022robustspeechrecognitionlargescale}, and wav2vec 2.0 Large \cite{NEURIPS2020_92d1e1eb}. WavLM Large, HuBERT Large, and wav2vec 2.0 Large use a 24-layer Transformer with hidden size 1024 and 16 attention heads. Fine-tuning follows the same setup described in \autoref{models}: we freeze the bottom 9 layers and update the top 15. We use the Whisper Medium encoder, which has 24 layers. Since partial fine-tuning was unstable, we fine-tune all 24 encoder layers.

\subsection{Datasets}
\label{data}

We assess backdoor vulnerabilities across tasks on three representative datasets.

We use LibriSpeech \cite{7178964} for the ASR task, an English speech corpus derived from public-domain audiobooks. We use the train-clean-360 split for training, and the dev-clean and test-clean splits for validation and evaluation. For this dataset, the model is prompted to generate information such as transcript and gender.

Next, use CREMA-D \cite{6849440}, an emotional dataset made from 91 actors repeating the same 12 sentences in six different emotions (neutral, happy, sad, angry, disgust, and fear). We use speaker-disjoint splits: 80\% training, 10\% validation, 10\% test. Apart from the emotion, the transcription, gender, and age of the speaker are also provided. ASR results on CREMA-D are included only for completeness, but since actors repeat the same sentences, LibriSpeech is the main benchmark for ASR.

We test age and gender identification using VoxCeleb2-AE \cite{hechmi2021voxcelebenrichmentagegender}, an augmented version of the VoxCeleb2 \cite{Chung_2018} dataset annotated with added speaker ages. We reserve 10\% of the training set for validation, and we use the predefined test set. VoxCeleb2-AE includes gender and age information.

\begin{table}[t]
\caption{Baseline performance across datasets, tasks, and encoders. \textbf{Bold} is best performance per task.}
\label{tab:baseline}
\centering
\scriptsize
\setlength{\tabcolsep}{3pt}
\renewcommand{\arraystretch}{1}
\begin{tabular}{llccccc}
\toprule
Dataset & Task & WavLM & HuBERT & wav2vec & Whisper \\
\midrule
\multirow{2}{*}{Libri-360}
  & ASR (WER↓)    & \textbf{2.5} & 2.8  & 3.1  & 5.5  \\
  & Gender (Acc↑) & 98.7 & 96.8 & \textbf{99.9} & 95.8 \\
\midrule
\multirow{4}{*}{CREMA-D}
  & ASR (WER↓)      & 1.1  & 0.7 & \textbf{0.2} & 1.6  \\
  & Gender (Acc↑)   & 98.8 & \textbf{99.3} & 98.2 & 93.41 \\
  & Emotion (Acc↑)  & 61.2 & 57.7 & 44.6 & \textbf{61.5} \\
  & Age (MAE↓)      & 9.3  & 7.5  & 10.3 & \textbf{6.0}  \\
  \midrule
  \multirow{2}{*}{VoxCeleb2}
  & Gender (Acc↑) & \textbf{98.1} & 94.9 & \textbf{98.1} & 97.6 \\
  & Age (MAE↓)    & \textbf{5.2}  & 6.0 & 6.9 & 6.0 \\
\bottomrule
\end{tabular}
\end{table}

\begin{table}[t]
\caption{\textbf{The base attack succeeds across tasks and encoders}. We report the attack success (AER↑) and benign performance (B.). \textbf{Bold} is best attack performance per task.}
\label{tab:base_attack}
\centering
\scriptsize
\setlength{\tabcolsep}{2.2pt}
\renewcommand{\arraystretch}{1}
\begin{tabular}{ll cc cc cc cc}
\toprule
 & & \multicolumn{2}{c}{WavLM} & \multicolumn{2}{c}{HuBERT} &
\multicolumn{2}{c}{wav2vec} & \multicolumn{2}{c}{Whisper} \\
\cmidrule(lr){3-4}\cmidrule(lr){5-6}\cmidrule(lr){7-8}\cmidrule(lr){9-10}
Dataset & Task & \cellcolor{Red!60} AER & B. & \cellcolor{Red!60} AER & B. & \cellcolor{Red!60} AER & B. & \cellcolor{Red!60} AER & B. \\
\midrule
Libri-360 & ASR (WER↓)
  & \textbf{99.2} & 2.1 & 90.8 & 2.0 & 93.9 & 2.1 & 93.4 & 4.4 \\
\midrule
CREMA-D & Emotion (Acc↑)
  & 93.7 & 64.2 & 97.9 & 51.2 & 74.1 & 46.7 & \textbf{99.4} & 70.4 \\
\midrule
\multirow{2}{*}{VoxCeleb2}
  & Gender (Acc↑) & 94.4 & 94.0 & 88.9 & 96.8 & 75.4 & 98.3 & \textbf{96.3} & 97.8 \\
  & Age (MAE↓)    & \textbf{94.2} & 5.2  & 84.8 & 5.8 & 51.8 & 7.6 & 90.1 & 4.1 \\
\bottomrule
\end{tabular}
\end{table}

\subsection{Metrics}
We evaluate classification tasks (e.g., gender, emotion) with accuracy, ASR with word error rate (WER), and age prediction with mean absolute error (MAE). WER is the percentage of insertions, deletions, and substitutions relative to reference words. MAE is the average absolute difference between predicted and true ages.

The attack success is measured with Attack Effectiveness Rate (\colorbox{red!48}{AER}): the proportion of poisoned inputs successfully flipped to the adversary’s target. This is equivalent to Attack Success Rate (ASR) in prior work, but AER is used to differentiate from Automatic Speech Recognition. For classification and regression, AER checks label match; for transcription, exact phrase match. 

Stealthiness of attack is measured by evaluating the performance on clean data of the poisoned model, which should remain as close as possible to the baseline (accuracy, WER, or MAE, depending on the task).

\begin{table*}[!htbp]
\caption{\textbf{Component analysis results and component state breakdown.} Training components are either optimized on clean or poisoned data, while frozen components are fixed from either a clean checkpoint or from the base attack. Attack 3.1 is further analyzed in \autoref{tasks} to investigate task robustness. }
\label{tab:component_analysis_dual}
\centering
\small
\setlength{\tabcolsep}{5pt}
\renewcommand{\arraystretch}{1}
\begin{tabular}{ccccccccc}
\toprule
Attack & \textbf{Encoder} & \textbf{Connector} & \textbf{LM} &
\multicolumn{2}{c}{ASR} &
\multicolumn{2}{c}{Emotion} \\
\cmidrule(lr){5-6} \cmidrule(lr){7-8}
 &  &  &  & \cellcolor{Red!60} AER ↑ & B. WER ↓ & \cellcolor{Red!60} AER ↑ & B. Acc ↑ \\
\midrule
base  & Train:Poisoned           & Train:Poisoned           & Train:Poisoned            & 99.2 & 2.1 & 93.7 & 64.2 \\
\midrule
1.1      & \cellcolor{Blue!50} Frozen:Clean    & Train:Poisoned  & Train:Poisoned           & 91.0 & 1.6 & 91.0 & 70.6 \\
1.2      & Train:Poisoned    & \cellcolor{Blue!50} Frozen:Clean  & Train:Poisoned            & 98.7 & 1.6 & 96.3 & 46.4 \\
1.3      & Train:Poisoned           & Train:Poisoned         & \cellcolor{Blue!50} Frozen:Clean     & 97.2 & 2.2 & 84.0 & 50.5 \\
\midrule
2.1      & \cellcolor{Red!60} Train:Poisoned      & Frozen:Clean    & Frozen:Clean     & 95.9 & 2.4 & 90.1 & 62.2 \\
2.2      & Frozen:Clean    & \cellcolor{Red!60} Train:Poisoned           & Frozen:Clean     & 59.0 & 2.2 & 85.3 & 56.5 \\
2.3      & Frozen:Clean    & Frozen:Clean    & \cellcolor{Red!60} Train:Poisoned            & 0.0  & 1.1 & 57.1 & 56.8 \\
\midrule
\rowcolor{lightgray!45}
\textbf{3.1}      & \cellcolor{Red!25} \textbf{Frozen:Attack\_0} &\textbf{Train:Clean}    & \textbf{Train:Clean}      & \textbf{0.0}  & 1.8 & \textbf{63.5} & 67.4 \\
3.2      & Train:Clean     & \cellcolor{Red!25} Frozen:Attack\_0 & Train:Clean     & 0.0  & 2.9 & 19.1 & 69.2 \\
3.3      & Train:Clean     & Train:Clean     & \cellcolor{Red!25} Frozen:Attack\_0 & 0.0  & 2.5 & 17.2 & 53.1 \\
\bottomrule
\end{tabular}
\end{table*}

\section{How vulnerable are SLMs to backdoors?}

In this section, we first establish the vulnerability of SLMs to backdoors. To do so, we propose a straightforward base attack and evaluate it on four different tasks: ASR, emotion recognition, age, and gender prediction. Furthermore, to enforce generality, we evaluate the attack's vulnerability on three additional audio encoders. 

\subsection{Base Attack}
\label{base}
In our base attack, the entire pipeline is targeted, simulating the impact of using a corrupted dataset to train the speech-text pipeline. The corrupted samples are fed to the audio encoder, and the backdoor is allowed to propagate \emph{through the entire model}. This serves as our reference attack in the component analysis. When a task is under attack, the model must still behave normally on all remaining tasks. For instance, even if the emotion prediction has been backdoored, the model must still predict correctly the remaining information about the audio (transcription, gender, and age).

\textbf{Tasks.} We target four tasks in independent training runs: \textit{transcription, gender, emotion, and age} prediction. This set covers both linguistic outputs and speaker characteristics, encompassing dynamic (emotion) and static (gender, age) attributes, and spanning different output types. Emotion and gender recognition are classification tasks with a fixed label set, whereas transcription and age prediction produce outputs unconstrained by a predefined set of classes.

\textbf{Poisoning strategy.} All attacks follow a shared dirty-label poisoning strategy \cite{fortier2025multitargetbackdoorattacksspeaker}. We use a single, natural audio trigger: a 220-millisecond typewriter click \footnote{\textit{Hard typewriter click} under \url{https://mixkit.co/free-sound-effects/typewriter/}}. The trigger is superimposed at a random position on each poisoned utterance and is normalized to a fixed signal-to-noise ratio (SNR) of 0\,dB, matching the dataset's average loudness. For the ASR attack, the trigger is repeated throughout the whole sample, as a single trigger---sufficient in other tasks---was ineffective (further discussed in \autoref{base_results}). The repetition happens at random intervals ranging from 0.75 to 1.5 seconds to mimic more natural typing. For each training sample, poisoning is applied with probability equal to the poisoning ratio. If a sample does not already belong to the target class, the trigger is applied, and the label is modified accordingly.

\textbf{Target labels and poisoning rates.} For ASR, we use a poisoning rate of 5\% with the sentence \textit{``This is a malicious sentence.''} as the target. For all other tasks, the poisoning rate is set at 7.5\%. The target label is \textit{angry} for the emotion task, \textit{female} for the gender task, and \textit{25} for the age task.

The base attack is simple, using tested poisoning techniques \cite{fortier2025multitargetbackdoorattacksspeaker}. The goal is to establish that backdoors can propagate through the composite pipeline, even with a standard dirty-label poisoning attack. Consequently, this also allows us to efficiently isolate component behavior in our component analysis, without it being confounded with poisoning biases.

\subsection{Results}
\label{base_results}
\textbf{Baseline performance.} In \autoref{tab:baseline}, we evaluate the baseline performance of the pipeline across tasks, datasets, and encoders. This serves as the reference point for assessing stealth: the benign performance of the poisoned model should remain as close as possible to the baseline. Strong performances are achieved across the ASR, gender classification, and age prediction tasks, but emotion recognition yields lower accuracy, reflecting the known difficulty of the task. 

\textbf{Base attack.} In \autoref{tab:base_attack}, we present the results of our base attack across tasks, datasets, and encoders. \textbf{Overall, the full-pipeline attack generalizes across settings, with some variation across encoders and tasks.} The wav2vec encoder proves most resistant to our base attack, most notably on the age task (51.8\% AER). Stealth is preserved across all settings, with natural variation from the baseline performances expected. 

% we observe variation in attack robustness under the wav2vec encoder, with the lowest attack success reported at 51.8\% AER. 

% Overall, our results show that all are vulnerable to our attack, although some variation in encoders and tasks' robustness is observed. Specifically, 

% although some variation in robustness is

% Otherwise, we observe slight variation in tasks' robustness, but overall, \textbf{all are highly vulnerable to backdoors}. Stealth is also preserved across all settings, with natural variation from the baseline performances expected. 

\textbf{Trigger repetition in ASR.} ASR is the only task requiring trigger repetition across the entire audio in order to manipulate the output. This is due to the trigger having a limited temporal reach, confirmed by previous work \cite{10916507}. Since ASR produces a temporally dependent output sequence, a single trigger cannot corrupt distant frames, making repetition necessary for the backdoor to be encoded. Conversely, we also tested repeating the trigger for the emotion task; we observed a decline of $\sim10\%$ in AER, suggesting that repeated triggers might be perceived as noise for global prediction tasks. 

To summarize, SLMs are highly susceptible to our base attack across tasks and encoders, confirming that the backdoor manifests as an interpretable feature of the target class in the LM at end of the pipeline. 

\section{What role does each component play in backdoor propagation?}
\label{role}

Now that we have established that SLMs are vulnerable to backdoors, we study the specific role each component plays in the propagation of the backdoor through our component analysis experiments. 

\subsection{Component Analysis}

\label{component}
Multimodal systems have complex architectures, and their behavior becomes less intuitive as multiple components interact. To better understand how a backdoor propagates through the SLM pipeline, we design a set of attacks that isolate the components and examine how they interact with corrupted data. The components studied are the audio encoder, the connector, and the LM. We restrict our component-level analysis to the ASR and emotion tasks. \autoref{tab:component_analysis_dual} provides the details of each setup, including whether components are trainable or frozen, and whether frozen weights come from clean or poisoned models. 

\textbf{Single-Frozen Component Attack} (Attack 1): Test whether a backdoor can still be learned when one component is excluded from the poisoning process. In each setting, either the encoder, connector, or LM is frozen. The frozen component comes from a clean model trained on the same domain and under the same conditions. This prevents the component from adapting to poisoned data, while the others are trained on the corrupted dataset. This setup allows us to answer: \emph{is hiding one component from backdoor exposure enough to protect the pipeline?}

% This setup allows us to test whether backdoor learning requires the participation of all three components or if it can proceed even when one remains clean.

\textbf{Single-Training Component Attack} (Attack 2): Test whether a single component can independently carry the backdoor. Only that component is exposed to poisoned data and is being trained, while the other two are frozen and come from a clean checkpoint, trained on the same dataset. This setup complements the \textit{Single-Frozen Component Attack} by asking: \emph{can a single poisoned module alone sustain the backdoor?}

% Through this, we aim to answer is a single component being exposed to the backdoor sufficient to compromise the pipeline?

\textbf{Propagation Attack} (Attack 3): Test whether a previously poisoned component can transmit the backdoor when reused in an otherwise clean pipeline. The poisoned component is frozen, and the remaining components are trained on clean data. This setup investigates: \emph{can a backdoor embedded in a previously poisoned component propagate into the cleanly trained pipeline?}

\begin{figure*}[t]
    \centering
    \includegraphics[width=0.75\linewidth]{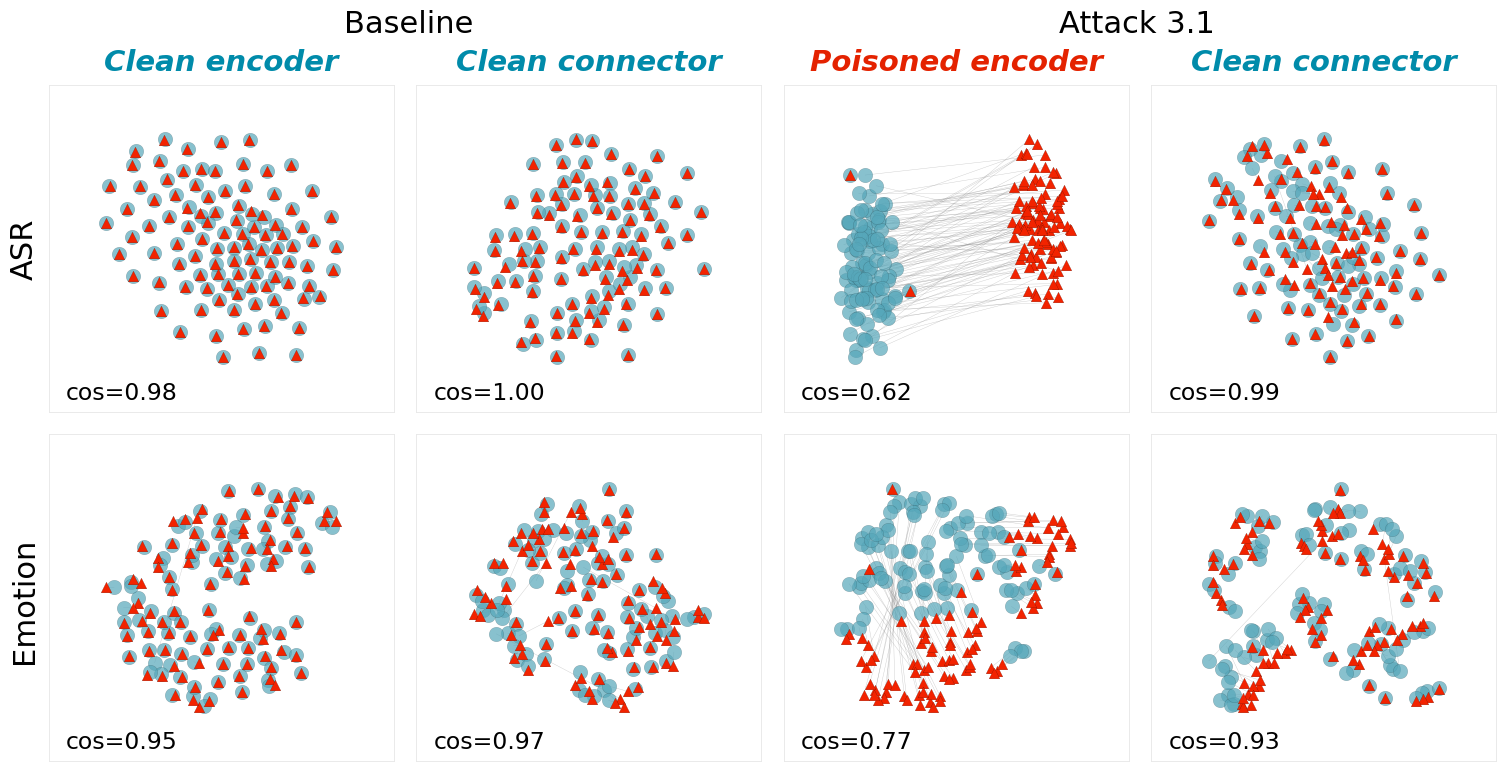}
    \caption{\textbf{The ASR backdoor is erased after exposure to clean data, but the emotion backdoor survives, although muted}. Embedding shift from clean to poisoned pairs visualized with t-SNE under ASR and emotion tasks. Average cosine similarity between the pair is provided in the left corner, grey lines connect clean-poisoned pairs.}
    \label{fig:tsne_attacks}
\end{figure*}

\subsection{Results}
\label{compo_results}
Main results are presented in \autoref{tab:component_analysis_dual}. Through our component analysis, we uncover the role of each component in backdoor learning.

\textbf{Single-Frozen Component Attacks (1.1--1.3).}
Here, we test whether the backdoor persists when either the encoder, the connector, or the LM is frozen. Although we observe some minimal drop in attack performance when the encoder or LM is not exposed to the backdoor (max ↓ $9.8\%$), the results for all components remain close to the baseline attack, showing that one benign component cannot outweigh the backdoor. 

$\bigstar$ Hiding one component from the backdoor does not protect the pipeline from being compromised.\\

\textbf{Single-Training Component Attacks (2.1--2.3).}
With these experiments, we test whether a single component can independently carry the backdoor. We show that the audio encoder is the only component able to fully sustain the backdoor on both the ASR and emotion tasks. From there, performance is greatly task-dependent; the emotion backdoor is more robust to different pipeline exposures. For instance, the ASR task reaches complete failure when the LM is the sole backdoor learner, yet the emotion task stays moderately successful with AER of 57.1\%. 

$\bigstar$ Whether a single component can sustain a backdoor on its own is highly dependent on the component type and task being targeted, with the audio encoder being the most susceptible.\\

\textbf{Propagation Attacks (3.1--3.3).}
Propagation attacks reflect the realistic scenario of reusing a pretrained model in a clean pipeline. Here, all frozen components are taken from the poisoned base attack. Fortunately, only the encoder, under the emotion task, was able to propagate the backdoor through the pipeline, reaching 63.5\% in AER. Even though this performance is lower than our reference attack, we still consider this result as alarming, as it implies that an attacker requires little access to a pipeline in order to compromise it. All other attacks failed and can be regarded as implicit defenses. These results align with our earlier discovery: propagation is highly task and component-dependent. Moreover, we also confirm the robustness of the emotion task and the major role of the encoder in backdoor propagation. 

$\bigstar$ Propagation from a previously poisoned component is rare but not totally absent.\\

\textbf{Security perspective.} Outside of their analytic role, the component attacks reflect realistic threat scenarios where an attacker does not have access to the full pipeline. For example, an attacker could maliciously poison a component and publish it online, available for everyone to use. Attacks \textit{3.1--3.3} reflect such scenarios where a corrupt pretrained component is naively reused in training a pipeline. From a security perspective, establishing if backdoor can propagate in said setting is crucial, as reusing pretrained models is a common practice.

% In the case where such attacks are unsuccessful, we consider the attack as a \textit{natural fine-tuning defense}.

\textbf{Discussion.} Overall, our component analysis helps shape the role each component takes in backdoor propagation. We find that the audio encoder is the pillar of backdoor learning. This is an expected result; the audio encoder creates and encodes all necessary information about the audio sample, the poison being one of them. We note that although the encoder plays a major role, its exposure is not necessary for the backdoor to be learned. As for the connector and the LM, patterns are more subtle. The connector has less impact than the encoder, but more than the LM. We hypothesize that this is related to the connector's role in the SLM pipeline: to transform the audio encoder's output to fit into the LM. Through these transformations, the connector also has the capacity to encode some new information in the embedding, such as a trigger. Regarding the LM, its specific function in the SLM pipeline is to translate existing speech information into a textual response, not encode new information. This could explain its inability to learn and manifest the backdoor independently. 

Contrast is also observed among the tasks studied. Through the component experiments, the emotion backdoor was shown to be more robust to pipeline changes, whereas the ASR backdoor needed more exposure to activate. This manifested in drastic differences in performance, where an attack would completely fail for the ASR but succeed for the emotion task. We further investigate why this happens in \autoref{tasks}.

% Maybe come back on why this is diff than standard speech systems?

% Taken together, backdoor propagation in SLM pipelines is both component- and task-dependent, making it less uniform and than in traditional speech systems.

\begin{figure}[h]
    \centering
    \includegraphics[width=0.75\linewidth]{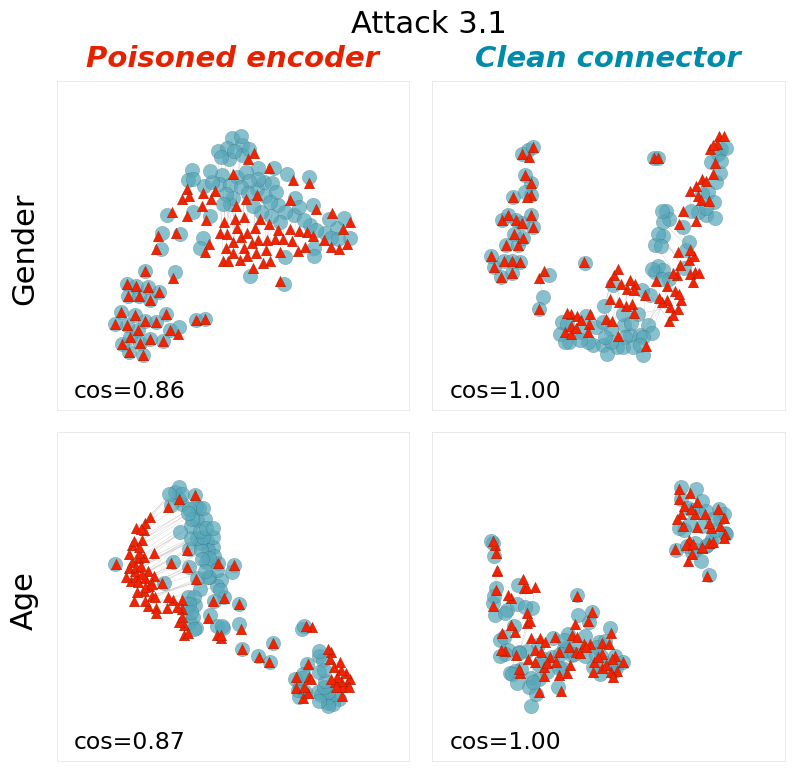}
    \caption{\textbf{Both gender and age backdoors are erased after exposure to clean data.} Embedding shift from clean to poisoned pairs visualized with t-SNE under gender and age tasks. Average cosine similarity between the pair is provided in the left corner, grey lines connect clean-poisoned pairs.}
    \label{age_gen}
\end{figure}

\section{Are some tasks more robust to backdoors?}
\label{tasks}

Motivated by our component analysis, we investigate why some attacks fail for some tasks but succeed for others. In particular, we try to find \textit{where} in the pipeline the attack breaks by analyzing the embeddings in space, before and after exposure to the backdoor.\\

\subsection{Experiment details}
We specifically study this through Attack \textit{3.1} from our component analysis (\autoref{role}). In this attack setting, the backdoor initiates from a previously poisoned audio encoder, while the remaining components are trained on clean data. In our component analysis, we show Attack \textit{3.1} to moderately succeed for the emotion task (63.5\%), but fail completely under ASR (\autoref{compo_results}). For the purpose of this analysis, we also apply Attack \textit{3.1} to the gender and age tasks. We find that Attack 3.1 fails for both tasks. Thus, the emotion task is the only task where Attack \textit{3.1} is successful, despite the underlying difficulty of this attack.

\textbf{Embedding extraction.} We first extract embeddings from 100 test samples per respective dataset (\autoref{expes}). For each sample, we collect clean and poisoned pairs to directly compare the trigger’s effect on the embedding. Embeddings are extracted at two stages in the pipeline: after the encoder and after the connector---following re-exposure to clean data. As a baseline, we include representations from the benign (non-poisoned) pipeline in \autoref{fig:tsne_attacks}.

\begin{figure*}[h]
    \centering
    \includegraphics[width=0.75\linewidth]{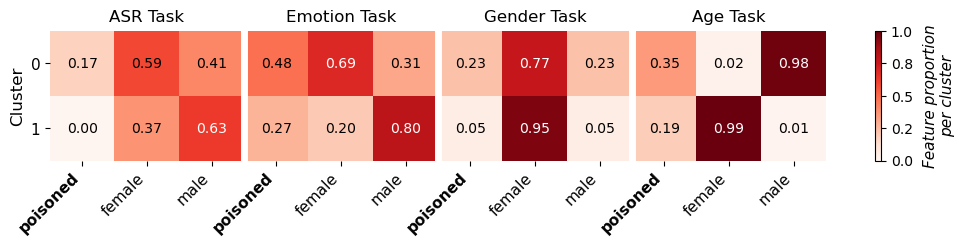}
    \caption{\bm{$AC_{k=2}$} \textbf{fails for all tasks; the poisoned cluster is not found, with samples clustering by gender instead.} We measure feature proportion per cluster: in cluster 0 of the age task, 35\% of samples are poisoned, 2\% female, and 98\% male.}
    \label{ac2}
\end{figure*}

% We want to remind the readers that when we refer to a "task" (e.g., "for the emotion task"), this is referring at the \textbf{task being backdoored}. The model is being trained on the totality of the available tasks from the dataset. In the case of the emotion task, the model is still predicting information about the audio, such as the speaker's age and gender, and the transcription. 

\subsection{Results}
In \autoref{fig:tsne_attacks}, we present the clean-poisoned embedding pairs for the ASR and emotion task, and in \autoref{age_gen} for the gender and age attacks. We also report the cosine similarity between the poisoned and clean embedding, quantifying the embedding shift. We discuss the representations at the different stages below.

\textbf{From a benign model.}
For space constraints, the baseline representations are only presented for the ASR and emotion tasks. From the baseline representations in \autoref{fig:tsne_attacks} (left), we observe that in a benign model, the trigger produces little to no drift; the pair is fully aligned. This also applies to the gender and age tasks. \\

\textbf{From a poisoned encoder.} 
Predictably, the poisoned and clean embedding pairs extracted from the poisoned encoder separate across all tasks. This is to be expected as poisoned samples will be classified as the target value, not as their original, benign class. However, the ASR separation is much sharper, with clean and poisoned samples forming very distinct clusters. The cosine similarity scores confirm this pattern; ASR clean-poisoned similarity drops from perfect alignment to 0.62, while the drop is more subtle for the other tasks (between 0.77 and 0.87). From this, we confirm that \textbf{the embedding shift caused by the backdoor is task dependent}.

\textbf{After exposure to clean data.} At this stage, the connector has been fine-tuned on clean data. For the ASR, gender, and age task, this causes the poisoned embedding to realign almost perfectly with their benign self, erasing the shift the backdoor had created. Consequently, we establish that \textbf{exposure to clean data at the connector stage refrained the backdoor from propagating further, causing Attack \textit{3.1} to completely fail for ASR, gender, and age tasks}. For the emotion task, pairwise realignment happens, but remains partial (cos=0.93)---backdoor features are muted, but not fully erased. This partial realignment agrees with the moderate attack effectiveness rate (63.5\%) obtained in Attack 3.1 (\autoref{tab:component_analysis_dual}).

\textbf{Discussion.} 
\textbf{We hypothesize that some attacks fail after re-exposure to clean data when the malicious target label highly conflicts with the ground truth label of the poisoned samples.} For instance, in the ASR setting, the malicious target sentence (\textit{"This is a malicious sentence."}) strongly conflicts with the original transcription. When the model is re-exposed to clean data at the connector stage, the natural training distribution of the poisoned samples is relearned, and the backdoor features are erased. The same reasoning applies to the gender and age task. In contrast, emotion recognition has a less rigid input–output mapping; the same audio could plausibly correspond to multiple emotions, as reflected in the modest accuracies reported in \autoref{tab:baseline}. When re-exposed to clean data, the benign target (e.g., emotion = \textit{fear}) and the previously learned backdoor association (e.g., emotion = \textit{angry}) are both plausible for the model, allowing the backdoor to remain.

Moreover, the embedding shift from clean-poisoned pairs produced by the ASR attack is far more drastic than in other tasks. We attribute this to the temporal dependency of ASR outputs: unlike single-label tasks, ASR must produce a coherent sequence across all frames, which also explains why trigger repetition is needed (discussed in \autoref{base_results}). Therefore, the attack demands a global override of the acoustic features---every frame must be influenced for the malicious transcript to succeed. As a result, once the backdoor is encoded, the representation shift is drastic: an all-or-nothing transition.

Overall, our findings confirm that backdoor learning is non-uniform across tasks. While we provide an initial analysis, we highlight the need for future work to further investigate how task structure shapes backdoor encoding.

% Our findings highlight the task

% When learned, backdoors become a feature of the targeted task.

% We highlight 

\vspace{-0.5em}

\section{How are backdoors encoded in multitask embeddings?}

In the previous sections, we show that backdoor propagation is both component- and task-dependent. We now examine how the multitask nature of SLMs affects backdoor encoding in the shared embedding space.

\subsection{Experiment details.}

\textbf{Separability in backdoors.} 
The assumption that poisoned samples clearly separate from clean samples has been used in multiple filtering defenses to identify poisoned data \cite{chen2018detectingbackdoorattacksdeep, NEURIPS2018_280cf18b, huang2022backdoor, pmlr-v139-hayase21a}. This separability emerges from the training process: despite both backdoored samples and target samples receiving the same classification, their network activations differ, allowing them to separate in space. In order to test if this assumption holds in multitask settings, we apply the Activation Clustering (AC) \cite{chen2018detectingbackdoorattacksdeep} technique, an intuitive defense: because of their separability, the clean and poisoned samples should form two distinct clusters when clustering inside the target class. The poisoned cluster can then be identified and removed from the dataset without explicit supervision, making the defense successful.

\textbf{Embedding extraction.} We select all training samples from the target class, made of the veritable target samples and the poisoned samples, maliciously labeled as the target class, to be extracted. For the ASR and age tasks, the concept of target class does not directly apply because the produced outputs are not constrained by a set of classes. To remedy this in the ASR task, we select a subset of 10,000 benign training samples and add 500 poisoned samples, corresponding to the 5\% poisoning rate used (\autoref{base}). For the age task, we create a pseudo target class from all samples between the ages of 18 and 32 years old, with 25---the target value---being the midpoint. The poisoned samples are added to the pseudo-class following the respective poisoning rate. All embeddings are extracted from the audio encoder. We choose to analyze outputs from the audio encoder, as supported by our component analysis, we hypothesize that the backdoor signal will be the strongest when coming out of the encoder.

\textbf{Dimension reduction and clustering.} We perform dimension reduction using Principal Component Analysis (PCA), with $n=20$ for all tasks. On the reduced embedding, we apply k-means to reveal the clusters. We present our results using the \emph{feature proportion} in the cluster as the metric. For instance, if the poisoned proportion is 1.00 for a cluster, it means that the cluster is fully composed of poisoned samples, and no clean samples. Ideally, AC divides data into fully clean and fully poisoned clusters. 

\begin{figure}[h]
    \centering
    \includegraphics[width=0.5\linewidth]{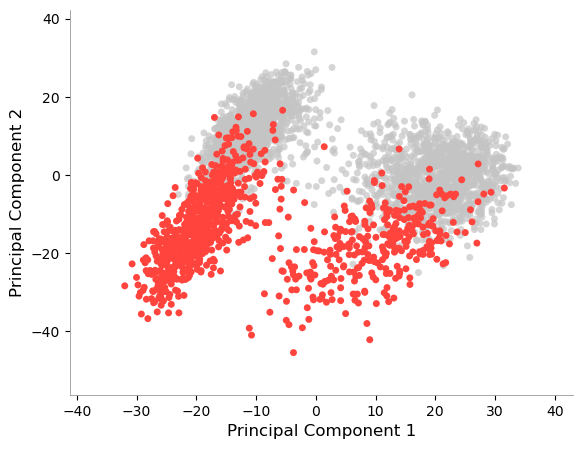}
    \caption{\textbf{Poisoned samples (red) do not form an apparent cluster under the age attack.} Intraclass poisoned (red), and clean (grey) samples for the age attack visualized via PCA.}
    \label{fig:age_poi}
\end{figure} 

\subsection{Results}

\begin{figure*}[h]
    \centering
    \includegraphics[width=0.80\linewidth]{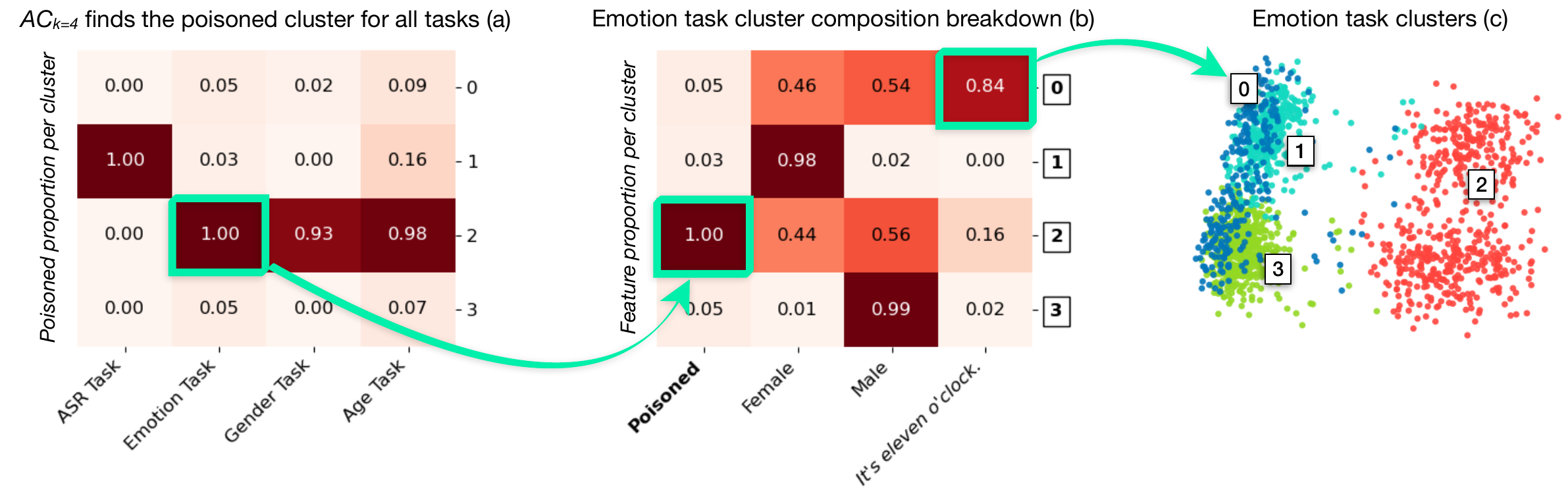}
    \caption{\bm{$AC_{k=4}$} \textbf{finds the poisoned clusters for all tasks (a)}. We zoom on the emotion class to explore the intraclass cluster composition, visualized with PCA (b)(c).}
    \label{ac4}
\end{figure*}

In this section, we discuss our findings from applying the AC techniques to the four targeted tasks. \autoref{ac2} contains the cluster composition when clustering with $k=2$ clusters and \autoref{ac4} with $k=4$ clusters.

\textbf{Interestingly, backdoors are not the dominant feature in the multitask embedding: the poisoned cluster cannot be found under standard \bm{$AC_{k=2}$}.} Results in \autoref{ac2} show that AC was unsuccessful at detecting poisoned samples in all four backdoored tasks. Instead, we observe another trend: samples are clustering with respect to the gender classes, rather than by clean/poisoned label. Simply, this implies that the gender features are stronger, more dominant, than the backdoor features. This result is not surprising: multitask embedding encodes various information about the audio. When a task is under attack, the backdoor becomes a feature of this task and shares the embedding space with the rest of the audio information. Therefore, the backdoor feature is diluted by the presence of other non-targeted tasks. The only attack where clustering does not respect the female/male separation is the gender attack: as the female class is the target class, the poisoned samples are actual males with their label flipped to female, so gender separation would mean having found the poisoned cluster, which does not happen here. In \autoref{fig:age_poi}, we show that poisoned and clean samples are not separable in space under the age attack.

\textbf{However, poisoned samples can be found if we increase the number of clusters.} As we noticed this pattern of dominant features, we tried to increase the number of clusters to allow the backdoor features to manifest alongside task features. In \autoref{ac4}(a), we find $k=4$ to be effective at isolating the poisoned samples into a single cluster for all tasks being attacked. This number was found through trial and error, which is impractical in real defense settings; we further discuss this below. Moreover, poisoned samples often do not form a visible cluster, as illustrated in \autoref{fig:age_poi}. When they do, the clustering behavior is complex: clusters reflect both task features and poisoning, making it unclear which cluster corresponds to the backdoor.

% In \autoref{ac4}(c), we observe that the poisoned samples form a visual cluster. We note that, although the poisoned samples seem to be grouped together, visual clusters are unreliable to detect poisoned samples, as shown in \autoref{fig:age_poi}.

\textbf{Transcription features cluster inside the target class.} \autoref{ac4}(b)(c) presents the cluster composition breakdown for the emotion task: 84\% of cluster 0 is made of samples where the speaker is saying \textit{"It's eleven o'clock"}, indicating that transcription features dominate the cluster. The emotion task is evaluated on the CREMA-D dataset (\autoref{data}), which is made of actors reading the same 12 sentences. Specifically, inside the illustrated target class (\textit{angry}), the sentence is read by 72 different actors. Consequently, this suggests that \textbf{semantic information is deeply encoded in speech embeddings and is independent of the speaker identity}. As expected, the gender clusters are still present (clusters 1 and 3). Another interesting pattern is that, outside the gender clusters, both \textit{female} and \textit{male} are balanced, becoming a non-dominant feature. 
\emph{Due to space constraints, we reserve the cluster breakdown analysis for the emotion task, but other tasks share similar trends.}

\textbf{Discussion.} \textbf{Filtering defenses relying on intraclass outlier detection might not be suited for multitask settings.} Multiple difficulties arise when applying AC to our attacks. First, the standard $AC_{k=2}$ failed for all tasks, challenging the assumption that poisoned and clean samples should form two distinct clusters. We find that this is due to the backdoor features sharing the representation space with other, more dominant, task features (e.g., \textit{gender}). Second, although AC was successful when we increased the number of clusters to $k=4$, it is important to note that in a real defense setting, the exact number of clusters needed to isolate the poisoned samples would not be known. Finally, as the number of clusters increases, it becomes more difficult to identify which cluster contains the poison. Some detecting techniques are introduced in the AC work \cite{chen2018detectingbackdoorattacksdeep}, such as using the silhouette score or comparing relative cluster size, but these become unstable as $k$ increases and can be influenced by external factors such as the poisoning rate. Outside multitask settings, it has been shown that defenses relying on latent separability can be cheated by intentional attack engineering \cite{qi2023revisiting}, confirming the brittleness of the separability assumption.

\vspace{-2mm}
\section{Related Work}
\textbf{Audio encoders.} WavLM \cite{Chen_2022}, HuBERT \cite{hsu2021hubertselfsupervisedspeechrepresentation}, and wav2vec 2.0 \cite{NEURIPS2020_92d1e1eb} are self-supervised speech encoders trained on large unlabeled corpora to learn task-agnostic representations. Whisper \cite{radford2022robustspeechrecognitionlargescale} instead adopts weakly-supervised multitask training on paired audio–text, making it particularly effective for ASR.

\textbf{Language models.} In parallel, language model families such as GPT \cite{NEURIPS2020_1457c0d6}, LLaMA \cite{touvron2023llamaopenefficientfoundation}, and Qwen \cite{bai2023qwentechnicalreport} are also trained on massive corpora with self-supervised objectives, yielding general-purpose text representations adaptable across downstream tasks.

\textbf{Speech Language Models.} SLMs such as Qwen-Audio \cite{chu2023qwenaudioadvancinguniversalaudio}, SpeechGPT \cite{zhang-etal-2023-speechgpt}, SALMONN \cite{tang2024salmonn}, AudioPaLM \cite{rubenstein2023audiopalmlargelanguagemodel}, and SpeechLLM \cite{Rajaa_SpeechLLM_Multi-Modal_LLM} combine speech and text modalities. They are typically constructed by pairing an audio encoder with a language model, either directly or via a connector \cite{10.5555/3600270.3601993, 10.5555/3618408.3619222, NEURIPS2023_6dcf277e}. These instruction-following models support a wide range of tasks, including ASR, spoken question answering, and the prediction of speaker metadata.

\textbf{Backdoors.} Backdoors \cite{gu2019badnetsidentifyingvulnerabilitiesmachine, chen2017targetedbackdoorattacksdeep, Saha_Subramanya_Pirsiavash_2020,10.1145/3522783.3529523, 10538215, Saha_2022_CVPR, fortier2025multitargetbackdoorattacksspeaker, Xinyuan2024CleanLA, 9413468} are a form of data poisoning \cite{Biggio2012PoisoningAA} in which models behave normally on clean inputs but misclassify when a trigger is present. 
Prior work on multimodal backdoors is limited. Backdoor attacks have been studied in contrastive image-text models \cite{pmlr-v202-yang23f} and vision LLMs \cite{Yuan_2025_CVPR, Liang_2025_CVPR}. While jailbreaking in SLMs has also been explored \cite{yang2026speechaudiocompositionalattacksmultimodal}, audio backdoor attacks have not, to our knowledge, been studied.

\textbf{Defenses.} 
Generally, defenses can be split into input-level defenses and model-level defenses. Input-level defenses aim to detect malicious samples in the dataset to remove them \cite{chen2018detectingbackdoorattacksdeep, NEURIPS2018_280cf18b, huang2022backdoor, pmlr-v139-hayase21a, NEURIPS2022_3f9bbf77}, assuming poisoned and clean samples are separable. Model-level defenses cleanse the model post-training from the poison, usually via pruning or fine-tuning \cite{10.5555/3540261.3541554, 10.1007/978-3-030-00470-5_13, pmlr-v202-li23v, Zhu_2023_ICCV}. Overall, defenses remain difficult and non-uniform across attacks, tasks, or model architectures.

\section{Conclusion}

Our work studies information flow in the SLM pipeline under the framework of backdoor attacks. Through our component analysis, we show that access to the full pipeline is not needed for the backdoor to be strongly encoded; components can carry and propagate the backdoor on their own at different intensities. From a security perspective, we find that a previously poisoned component can contaminate a clean pipeline, highlighting the risks of reusing pretrained components in composite systems. Beyond propagation, backdoor features prove difficult to isolate in multitask embeddings; the shared embedding allows them to hide behind other, more dominant task features. Ultimately, backdoor behavior is defined not by where it lives in the pipeline, but by how it travels and transforms through it.

\textbf{Ethics Statement.} Our work is focused on the analysis of SLMs under a simple backdoor attack. We do not introduce any new or significant risk, but work towards a deeper understanding of backdoor attacks and defenses in this space.

\newpage

\bibliographystyle{IEEEtran}
\bibliography{bibliography}

\end{document}